\documentclass[runningheads]{llncs}
\usepackage{graphicx}
\usepackage{amsmath,amssymb} % define this before the line numbering.
\usepackage{color}
\usepackage[width=122mm,left=12mm,paperwidth=146mm,height=193mm,top=12mm,paperheight=217mm]{geometry}
\usepackage{times}
\usepackage{epsfig}
\usepackage{algorithm}
\usepackage[noend]{algpseudocode}
\usepackage{mathptmx}
\usepackage{xfrac}
\usepackage{enumitem}
\usepackage{cite}
\usepackage{indentfirst}
\usepackage{booktabs}
\usepackage[scaled=0.81]{DejaVuSansMono}
\usepackage{url}
\usepackage{siunitx}
\usepackage{array}
\usepackage{makecell}

\expandafter\def\expandafter\UrlBreaks\expandafter{\UrlBreaks%  save the current one
  \do\a\do\b\do\c\do\d\do\e\do\f\do\g\do\h\do\i\do\j%
  \do\k\do\l\do\m\do\n\do\o\do\p\do\q\do\r\do\s\do\t%
  \do\u\do\v\do\w\do\x\do\y\do\z\do\A\do\B\do\C\do\D%
  \do\E\do\F\do\G\do\H\do\I\do\J\do\K\do\L\do\M\do\N%
  \do\O\do\P\do\Q\do\R\do\S\do\T\do\U\do\V\do\W\do\X%
  \do\Y\do\Z}

\begin{document}
% \linenumbers
\pagestyle{headings}
\mainmatter

\title{Value-aware Quantization \\for Training and Inference of Neural Networks} % Replace with your title

\titlerunning{Value-aware Quantization for Training and Inference of Neural Networks}

\authorrunning{Eunhyeok Park, Sungjoo Yoo, Peter Vajda}

\author{Eunhyeok Park \inst{1} , Sungjoo Yoo \inst{1} , Peter Vajda \inst{2} }

%Please write out author names in full in the paper, i.e. full given and family names. 
%If any authors have names that can be parsed into FirstName LastName in multiple ways, please include the correct parsing, in a comment to the volume editors:
%\index{Lastnames, Firstnames}
%(Do not uncomment it, because you may introduce extra index items if you do that...)

\institute{ Department of Computer Science and Engineering\\
	Seoul National University\\
	\email{ \{eunhyeok.park,sungjoo.yoo\}@gmail.com}
	\and
   Mobile Vision \\
   Facebook
   \\ vajdap@fb.com
}

%\author[shortname]{author1 \inst{1} \and author2 \inst{2}}
%\institute[shortinst]{\inst{1} affiliation for author1 \and %
%                      \inst{2} affiliation for author2}

\maketitle

\begin{abstract}
We propose a novel value-aware quantization which applies aggressively reduced precision to the majority of data while separately handling a small amount of large data in high precision, which reduces total quantization errors under very low precision. We present new techniques to apply the proposed quantization to training and inference. The experiments show that our method with 3-bit activations (with 2\% of large ones) can give the same training accuracy as full-precision one while offering significant (41.6\% and 53.7\%) reductions in the memory cost of activations in ResNet-152 and Inception-v3 compared with the state-of-the-art method. Our experiments also show that deep networks such as Inception-v3, ResNet-101 and DenseNet-121 can be quantized for inference with 4-bit weights and activations (with 1\% 16-bit data) within 1\% top-1 accuracy drop.
\end{abstract}

\section{Introduction}
 
As neural networks are being widely applied to server and edge computing, 
both training and inference need to become more and more efficient in terms of runtime, energy consumption and memory cost. 
On both servers and edge devices, it is critical to reduce computation cost in order to enable fast training, e.g., on-line training of neural networks for click prediction~\cite{Facebook1, Facebook2}, and fast inference, e.g., click prediction at less than 10ms latency constraints~\cite{TPU} and real-time video processing at 30 frames per second~\cite{RealtimeAI}. 
Reducing computation cost is also beneficial to reducing energy consumption in those systems since the energy consumption of GPU is mostly proportional to runtime~\cite{GPUEnergy}. 

Especially, training is constrained by the memory capacity of GPU. The large batch of a deep and wide model requires large memory during training. For instance, training of a neural network for vision tasks on high-end smartphones or self-driving cars having 4K images requires 24MB only for the input to the first layer of the network. During training, we need to store the activations of all the intermediate layers. Considering that the size of activation, at each intermediate layer, is comparable to that of input, the required memory size (batch size x total activation size of the network) can easily exceed the memory capacity of state-of-the-art GPU.

Reduced precision has potential to resolve the problems of runtime, energy consumption and memory cost by reducing the data size thereby enabling more parallel and energy-efficient computation, e.g., four int8 operations instead of a single fp32 operation, at a smaller memory footprint.
The state-of-the-art techniques of quantization are 16-bit training\cite{NVIDIA16} and 8-bit inference\cite{NVIDIA8}.
Considering the trend of ever-increasing demand for training and inference on both servers and edge devices,
further optimizations in quantization, e.g., 4 bits, will be more and more required.

In this paper, we propose a novel quantization method based on the fact that the distributions of weights and activations have the majority of data concentrated in narrow regions while having a small number of large data scattered in large regions. By exploiting the fact, we apply reduced precision only to the narrow regions thereby reducing quantization errors for the majority of data while separately handling large data in high precision. For very deep networks such as ResNet-152 and DenseNet-201, our proposed quantization method enables training with 3-bit activations (2\% large data). Our method also offers low-precision inference with 4 to 5-bit weights and activations (1\% large data) even for optimized networks such as SqueezeNet-1.1 and MobileNet-v2 as well as deeper networks.

\section{Related Work}

Recently, there have been presented several methods of memory-efficient training. In \cite{Sublinear}, Chen et al. propose a checkpointing method of storing intermediate activations of some layers to reduce memory cost of storing activations and re-calculating the other activations during back-propagation. In \cite{Reversible}, Gomez et al. present a reversible network which, during back-propagation, re-computes input activations utilizing output activations thereby minimizing the storage of intermediate activations.
The existing methods of checkpointing and reversible network are effective in reducing memory cost. However, they have a common critical limitation, the additional computation to re-compute activations during back-propagation. Considering that computation cost determines runtime and energy consumption of training on GPUs, the additional computation cost needs to be minimized.
As will be explained in the experiments, our proposed quantization method gives much smaller cost in both memory and computation than the state-of-the-art ones. More importantly, it has a potential of offering less computation cost than the conventional training method.

The state-of-the-art quantization methods of training and inference for deep networks, e.g., ResNet-152 are 16-bit training \cite{NVIDIA16} and 8-bit inference \cite{NVIDIA8}. In \cite{NVIDIA16}, Ginsburg et al. propose 16-bit training based on loss scaling (for small activations or local gradients) and fp32 accumulation.  
In \cite{NVIDIA8}, Migacz proposes utilizing Kullback-Leibler (KL) divergence in determining the linear range to apply 8-bit quantization with clipping.
There are studies towards more aggressive quantization for training, e.g., \cite{DoReFa, HALP}. In \cite{HALP}, De Sa et al. propose bit centering to exploit the fact that gradients tend to get smaller as training continues. However, these aggressive methods are limited to small networks and do not preserve full-precision accuracy for very deep models such as ResNet-152.

We classify quantization methods for inference into two types, linear and non-linear ones. The linear methods utilize uniform spacing between quantization levels, thereby being more hardware friendly, while the non-linear ones have non-uniform spacing mostly based on clustering. As the simplest form of linear quantization, 
in \cite{Rastegari}, Rastegari et al. show that a weight binarization of AlexNet does not lose accuracy. 
In \cite{Hubara}, Hubara et al. propose a multi-bit linear quantization method to offer a trade-off between computation cost and accuracy. In \cite{DoReFa}, Zhou et al. propose a multi-bit quantization which truncates activations to reduce quantization errors for the majority of data.

In \cite{Miyashita}, Miyashita et al. propose logarithm-based quantization and show that AlexNet can be quantized with 4-bit weights and 5-bit activations at 1.7\% additional loss of top-5 accuracy.
In \cite{Zhu}, Zhu et al. show that deep models can be quantized with separately scaled ternary weights while utilizing full-precision activations. In \cite{Park}, Park et al. propose a clustering method based on weighted entropy and show 5-bit weight and 6-bit activation can be applied to deep models such as ResNet-101 at less than 1\% additional loss of top-5 accuracy. In \cite{BQ}, Zhou et al. propose a clustering method called balanced quantization which tries to balance the frequency of each cluster thereby improving the utility of quantization levels.
Recently, several studies report that increasing the number of channels \cite{Channel} and adopting teacher-student models \cite{TS,ICLR} help to reduce the accuracy loss due to quantization. These methods can be utilized together with our proposed quantization method.

Compared with the existing quantization methods, our proposed method, which is a linear method, enables smaller bitwidth, effectively 4 bits for inference in very deep networks such as ResNet-101 and DenseNet-121 for which there is no report of accurate 4-bit quantization in the existing works.
%%%%%

\section{Motivation}

\begin{figure}
    \centering
    \includegraphics[width=\columnwidth]{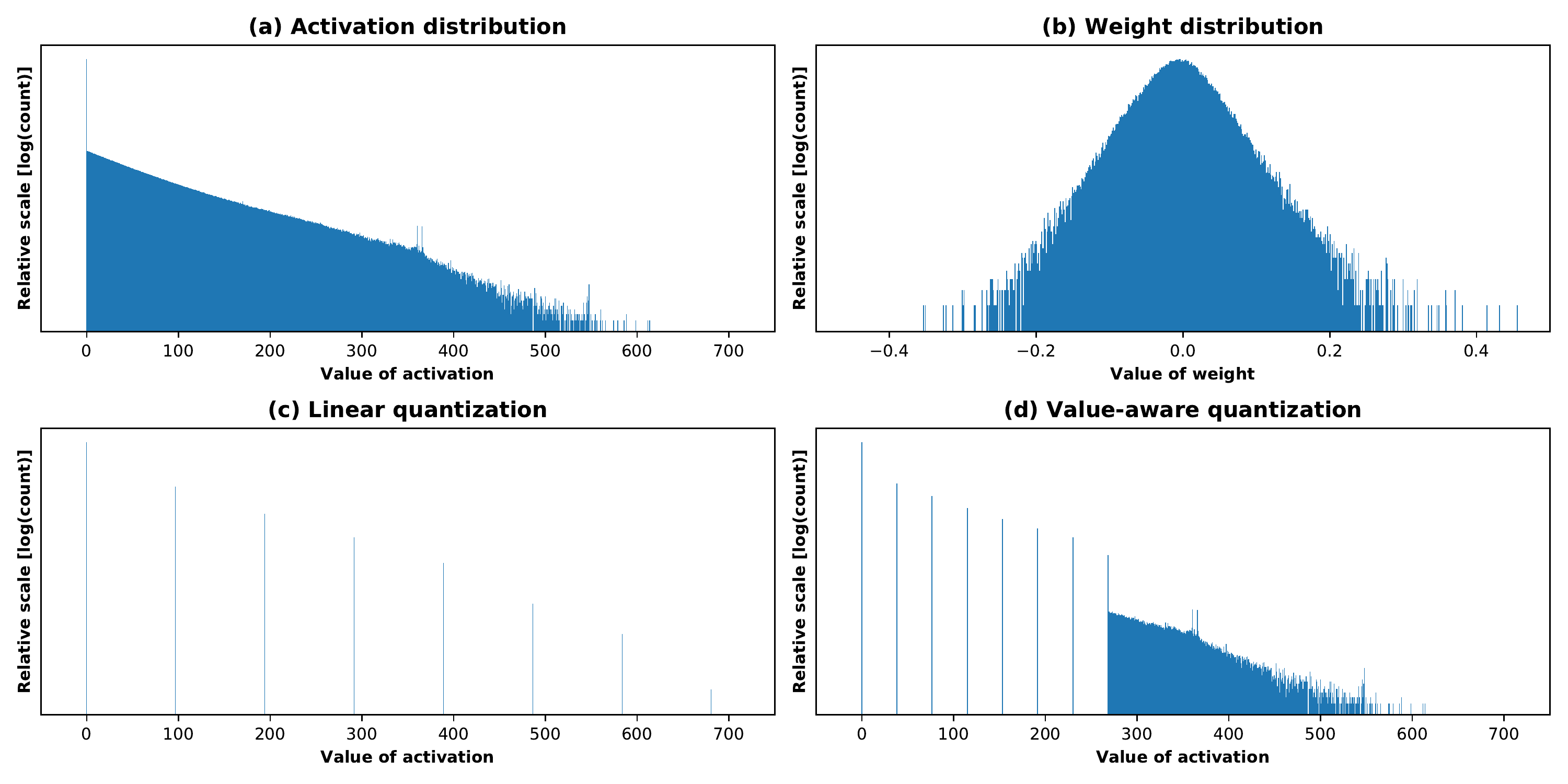}
    \caption{Activation and weight distributions of second convolutional layer in GoogLeNet.}
    \label{fig:1}
\end{figure}

Figure \ref{fig:1} (a) and (b) illustrate the distributions (y-axis in log scale) of activations and weights in the second convolutional layer of GoogLeNet. As the figures show, both distributions are wide due to a small number of large data. Given a bitwidth for low precision, e.g., 3 bits, the wider the distribution is, the larger quantization errors we obtain. Figure \ref{fig:1} (c) exemplifies the conventional 3-bit linear quantization applied to the distribution of activations in Figure \ref{fig:1} (a). As the figure shows, the spacing between quantization levels (vertical bars) is large due to the wide distribution, which incurs large quantization errors.

When comparing Figure \ref{fig:1} (a) and (c), it is clear that the majority of quantization levels is not fully utilized. Especially, the levels assigned to large values have much fewer data than those assigned to small values, which motivates our idea. 
Figure \ref{fig:1} (d) illustrates our idea. We propose applying low precision only to small data, i.e., the majority of data, not all. As the figure shows, the spacing between quantizaiton levels gets much smaller than that in the conventional linear quantization in Figure \ref{fig:1} (c). 
Such a small spacing can significantly reduce quantization error for the majority of data.
Large data have the larger impact on the quality of network output. Thus, we propose handling the remaining large data in high precision, e.g., in 32 or 16 bits. The computation and memory overhead of handling high-precision data is small because their frequency, which is called the ratio of large activations, in short, {\it activation ratio} (AR), is small, e.g., 1-3\% of total activation data.\footnote{We use two ratios of large data, one for large weights and the other for large activations. We use AR to denote the ratio of large activations.}

\section{Proposed Method}

Our basic approach is first to perform value profiling to identify large data during training and inference. Then, we apply reduced precision to the majority of data, i.e. small ones while keeping high precision for the large data. We call this method value-aware quantization (V-Quant).

We apply V-Quant to training to reduce the memory cost of activations. We also apply it to inference to reduce the bitwidth of weights and activations of the trained neural network.
To do that, we address new problems as follows.  

\begin{itemize}
    \item (Sections \ref{QuantBP} and \ref{FullLoss})
    In order to prevent the quality degradation of training results due to quantization, we propose a novel scheme called {\it quantized activation back-propagation}, in short, quantized back-propagation. We apply our quantization only to the activations used in the backward pass of training and perform forward pass with full-precision activations.
    \item (Sections \ref{LocalSorting} and \ref{QuantInf})
    Identifying large values requires sorting which is expensive. In order to avoid the overhead of global communication between GPUs for sorting during training, we propose performing sorting and identifying large values locally on each GPU. 
    \item (Sections \ref{RVQuant} and \ref{Annealing})
    We present new methods for further reduction in memory cost of training. In order to reduce the overhead of mask information required for ReLU function during back-propagation, we propose ReLU and value-aware quantization. For further reduction in memory cost, we also propose exploiting the fact that, as training continues, the less amount of large activations is required.
\end{itemize}

\subsection{Quantized Back-Propagation}
\label{QuantBP}

\begin{figure}
    \centering
    \includegraphics[width=\columnwidth]{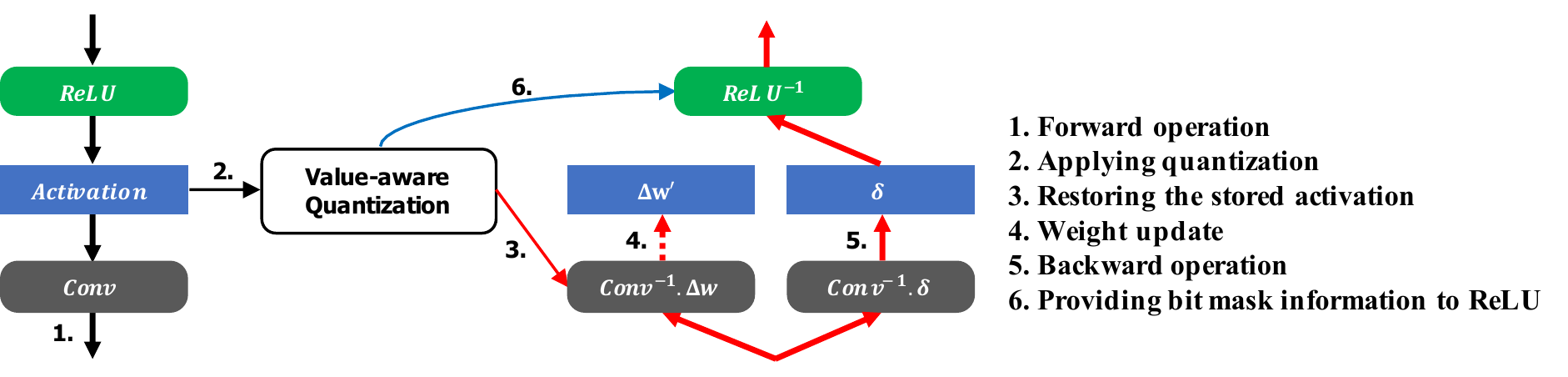}
    \caption{Value-aware quantization in training pipeline.}
    \label{fig:train}
\end{figure}

Figure \ref{fig:train} shows how to integrate the proposed method with the existing training pipeline.
As the figure shows, we add a new component of value-aware quantization to the existing training flow. In the figure, thick arrows represent the flow of full-precision activations (in black) and gradients (in red).

First, we perform the forward pass with full-precision activations and weights, which gives the same loss as that of the existing full-precision forward pass (step 1 in the figure).
During the forward pass, after obtaining the output activations of each layer, e.g., layer $l$, the next layer (layer $l+1$) of network takes as input the full-precision activations. 
Then, we apply our quantization method to them (those of layer $l$) in order to reduce their size (step 2).
As the result of the forward pass, we obtain the loss and the quantized activations.

During the backward pass, 
when the activations of a layer are required for weight update, 
we convert the quantized, mostly low-precision, activations, which are stored in the forward pass, into full-precision ones (step 3).
Note that this step only converts the data type from low to high precision, e.g., from 3 to 32 bits.
Then, we perform weight update with back-propagated error (red thick arrow) and the activations (step 4).

Note that there is no modification in the computation of the existing forward and backward passes.
Especially, as will be explained in the next subsection, when ReLU is used as activation function, the backward error propagation (step 5 in the figure) keeps full-precision accuracy.
The added component of value-aware quantization performs conversions between full-precision and reduced-precision activations and compresses a small number of remaining large high-precision activations, which are sparse, utilizing a conventional sparse data representation, e.g., compressed sparse row (CSR).

The conversion from full to reduced precision (step 2) reduces memory cost while that from reduced to full precision  (step 3) changes data type back to full precision one thereby increasing memory cost back to that of full precision. Note that the full-precision activations, obtained from the quantized ones, are discarded after weight update for their associated layer. Thus, we need memory resource for the stored quantized activations of the entire network and the full-precision input/output activations of only one layer, which we call working activations, for the forward/backward computation.

As will be explained later in this section, for further reduction in memory cost, the ReLU function consults the value-aware quantization component for the mask information which is required to determine to which neuron to back-propagate the error (step 6).

\subsection{Back-Propagation of Full-Precision Loss}
\label{FullLoss}

Our proposed method can suffer from quantization error in weight update since we utilize quantized activations. We try to reduce the quantization error by applying reduced precision only to narrow regions having the majority of data while separately handling the large data in high precision. 

Moreover, in state-of-the-art networks where ReLU is utilized as activation function, the back-propagated error is not affected by our quantization of activations as is explained below. Equation (1) shows how we calculate weight update during back-propagation for a multilayer perceptron (MLP). 

\begin{equation}
\Delta w_{ji} = \eta \delta_j y_i    
\end{equation}

\noindent
where 
$\Delta w_{ji}$ represents the update of weight from neuron $i$ (of layer $l$) to neuron $j$ (of layer $l+1$), $\eta$ learning rate, $\delta_j$ the local gradient of neuron $j$ (back-propagated error to this neuron), and $y_i$ the activation of neuron $i$. Equation (1) shows that the quantization error of activation $y_i$ can affect the weight update.
In order to reduce the quantization error in Equation (1), we apply V-Quant to activations $y_i$.

The local gradient $\delta_j$ is calculated as follows.

\begin{equation}
\delta_j =\varphi'(v_j)\Sigma_k \delta_k w_{kj}
\end{equation}

\noindent
where
$\varphi'()$ represents the derivative of activation function, $v_j$ the input to neuron $j$ and $w_{kj}$ the weight between neuron $j$ (of layer $l+1$) to neuron $k$ (of layer $l+2$). 
Equation (2) shows that the local gradient is a function of the input to neuron, $v_j$ which is the weighted sum of activations. However, if ReLU is used as the activation function, then $\varphi'()$ becomes 1 yielding $\delta_j =\varphi'(v_j)\Sigma_k \delta_k w_{kj}=\Sigma_k \delta_k w_{kj}$, which means {\it the local gradient becomes independent of activations}.
Thus, aggressive quantizations of intermediate activations, e.g., 3-bit activations can hurt only the weight update in Equation (1), not the local gradient in Equation (2). 
This is the main reason why our proposed method can offer full-precision training accuracy even under aggressive quantization of intermediate activations as will be shown in the experiments.

\subsection{Potential of Further Reduction in Computation Cost}
\label{potential}
Compared with the existing methods of low memory cost in training \cite{Sublinear}\cite{Reversible}, our proposed method reduces computation cost by avoiding re-computation during back-propagation. 
More importantly, our proposed method has a potential of further reduction in computation cost especially in Equation (1). It is because the activation $y_i$ is mostly in low precision in our method. Thus, utilizing the capability of 8-bit multiplication on GPUs, our method can transform a single 16-bit x 16-bit multiplication in Equation (1) into an 8-bit x 16-bit multiplication. In state-of-the-art GPUs, 
we can perform two 8-bit x 16-bit multiplications at the same computation cost, i.e., execution cycle, of one 16-bit x 16-bit multiplication, which means our proposed method can double the performance of Equation (1) on the existing GPUs.

Assuming that the forward pass takes $M$ multiplications, the backward pass takes 2$M$ multiplications while each of Equations (1) and (2) taking $M$ multiplications, respectively. Thus, the 2x improvement in computation cost of Equation (1) can reduce by up to 1/6 total computation cost of training. In order to realize the potential, further study is needed to prove that our proposed method enables 8-bit low-precision activations (with a small number of 16-bit high-precision activations) without losing the accuracy of 16-bit training \cite{NVIDIA16}.

Although our method can currently reduce computation cost utilizing only 8-bit multiplications on GPUs, 
its reduced-precision computation, e.g., 3-bit multiplications, offers opportunities of further reduction in computation cost for training in future hardware platforms supporting aggressively low precision, e.g., \cite{XilinxFPGA}.
%%%%%
\subsection{Local Sorting in Data Parallel Training}
\label{LocalSorting}

V-Quant requires sorting activations. Assuming that we adopt data parallelism in multi-GPU training, the sorting can incur significant overhead in training runtime since it requires exchanging the activations of each layer between GPUs. What is worse, in reality, such a communication is not easily supported in some training environments, e.g., PyTorch. 
In order to address the problem of activation exchange, 
we propose performing sorting locally on each GPU, which eliminates inter-GPU communication for activation exchange.
Then, each GPU performs V-Quant locally by applying the same AR, i.e., the same ratio of large activations.
Compared with the global solution that collects all the activations and applies the AR to the global distribution of activations, the proposed local solution can lose accuracy in selecting large values.
However, our experiments show that the proposed method of local sorting works well, which means that the selection of large values does not need to be accurate.

\subsection{ReLU and Value-aware Quantization (RV-Quant)}
\label{RVQuant}
% 아래 문단 내용에 약간 설명을 추가해봤습니다. 
%The ReLU propagates the error through the neuron having the positive activation. However, when we adopt V-Quant, some positive activation becomes zero due to the quantization thus the error can't be propagated through the in-place ReLU function. According to our observation, this degrades the network accuracy significantly. To prevent this problem, we need a bit mask, i.e., 1-bit memory cost for each neuron.

The error is back-propagated through the neurons the output activations of which are non-zero. When ReLU is adopted as activation function, the output activations often become zero. In such a case, in order to identify which neurons to propagate errors to, 
we need a bit mask, i.e., 1-bit memory cost for a neuron.
In case that the activations are quantized at a very small number of bits, e.g., 3 bits, the overhead of the bit mask is significant, e.g., one additional bit for 3-bit activation on each neuron. In order to reduce the overhead of mask information, we exploit the fact that each neuron needs to have either the output activation (for weight update) or the mask information (to block error back-propagation), not both at the same time.
Thus, given $K$ bits for low precision, we allocate one of $2^K$ quantization levels to the mask information while representing the activation value with $2^K - 1$ levels. We call this quantization ReLU and value-aware quantization (RV-Quant). As will be shown in the experiments, RV-Quant removes the overhead of bit mask while keeping training accuracy.

\subsection{Activation Annealing}
\label{Annealing}
According to our investigation, the required amount of large activations varies across training phases. To be specific, the early stage of training tends to require more large activations while the later stage tends to need less large activations. 
We propose exploiting the fact and adjusting AR in a gradual manner from large to small AR across training phases, which we call {\it activation annealing}. As will be shown in the experiments, activation annealing can maintain training quality while reducing the average memory cost across the entire training phases.

\subsection{Quantized Inference}

\label{QuantInf}

In order to obtain quantized neural networks for inference, we perform V-Quant as a post-processing of training, i.e., we apply V-Quant to the weights and activations of trained networks.
In order to recover from the accuracy loss due to quantization, we perform fine-tuning as follows.
We perform forward pass while utilizing the quantized network, i.e., applying V-Quant to weights and activations.
During back-propagation, we update full-precision weights.
As will be shown in the experiments, the fine-tuning incurs a very small overhead in training time, i.e., only a few additional epochs of training.
Note that we apply local sorting in Section \ref{LocalSorting} to avoid communication overhead when multiple GPUs are utilized in fine-tuning.

During fine-tuning, we evaluate candidate ratios for large weights and activations and, among those candidates, select the best configuration which minimizes the bitwidth while meeting accuracy requirements. Note that, as will be explained in the experiments, the total number of candidate combinations is small. 

In order to identify large activations meeting the AR, we need to sort activations, which can be expensive in inference. In order to avoid the sorting overhead, we need low-cost sorting solutions, e.g., sampling activations to obtain an approximate distribution of activations. Detailed implementations of quantized models including the low-cost sorting are beyond the scope of this paper and left for further study.
%If the activation is smaller than the threshold, we consider it as small one and apply low precision during inference.
%As in \cite{ISCA}, we obtain per-layer threshold using the validation set during the fine-tuning.
%Thus, as the results of fine-tuning, we obtain  weights in low and high precision, and per-layer activation thresholds.

%When the underlying hardware platform supports reduced precision operations, e.g., multiplication and addition in 8 or 4 bits, the quantized network can produce the output of each layer in three steps as follows.
%First, the reduced precision computation is performed with low-precision weights and input activations, which gives partial sums. Second, the large weights are computed with all the input activations and the results are accumulated to the partial sums. Third, the large input activations are computed with all the weights to accumulate the results to the partial sums, which gives the final results.

%Compared with the total computation cost of full-precision operations, our method can give significant reduction because the majority of operations are performed in reduced precision while a very small amount of computation is performed in mixed or full precision, e.g., a multiplication of low-precision weight and high-precision activation. We report the comparison of computation cost in the experiments.

\section{Experiments}

We evaluate our proposed method on ImageNet classification networks,
AlexNet, VGG-16, SqueezeNet-1.1, MobileNet-v2, Inception-v3, ResNet-18/50/101/152 and\\ DenseNet-121/201. We test the trained/quantized networks with ILSVRC2012 validation set (50k images) utilizing a single center crop of 256x256 resized image. 
We also use an LSTM for word-level language modeling \cite{LSTM1, LSTM2, LSTM3}.
We implemented our method on PyTorch framework \cite{PyTorch} and use the training data at Torchvision \cite{Torchvision}. 

The initial learning rate is set to 0.1 (ResNet-18/50/152 and DenseNet-201), or 0.01 (AlexNet and VGG-16). The learning rate is decreased by 10x at every multiple of 30 epochs and the training stops at 90 epochs. In SqueezeNet-1.1, MobileNet-v2 and Inception-v3, we use the same parameters in the papers except that we use a mini-batch of 256 and SGD instead of RMSprop. In addition, we replace ReLU6 in MobileNet-v2 with ReLU to apply V-Quant.

We apply V-Quant and RV-Quant to training to minimize memory cost. During training, in order to compress the sparse large activations on GPU, we use the existing work in \cite{Efficient}. In order to obtain quantized networks for inference, we perform fine-tuning with V-Quant for a small number of additional epochs, e.g., 1-3 epochs after total 90 epochs of original training.

We compare classification accuracy between full-precision models and those under RV-Quant (training) and V-Quant (training/inference). For each network, we use the same randomly initialized condition and perform training for different RV-Quant and V-Quant configurations.

\subsection{Training Results}

Table \ref{sweep} shows top-1/top-5 accuracy of ResNet-50 obtained, under V-Quant, varying the bitwidth of low-precision activation and the ratio of large activation, AR. The table shows that the configuration of 3-bit activations with the AR of 2\% (in bold) gives training results equivalent to the full-precision (32-bit) training in terms of top-1 accuracy, which corresponds to 6.1X (=1/((3+1)/32 + 0.04)) reduction in the memory cost of stored activation at the same quality of training.\footnote{Note that V-Quant still requires 1-bit mask information for each neuron. In addition, the sparse data representation of large data, e.g., CSR doubles the size of the original sparse data yielding the memory cost of 4\% with the AR of 2\%.} The table also shows that a very aggressive quantization of 2-bit activation and 1\% AR loses only 0.264\%/0.246\% in top-1/top-5 accuracy, which is comparable to the case of 5-bit quantization without large data (5-bit with AR 0\% in the table). 

Note that the total memory cost of activations includes that of stored activations of the entire network and that of full-precision working activations (input to the associated layer) required for weight update. Thus, the above-mentioned reduction of 6.1X is only for the memory cost of stored activations. We will give the comparison of total memory cost of activations later in this section.

\begin{table}[ht]
\small
\centering
\resizebox{\columnwidth}{!}{%
\begin{tabular}{ccccccc}
\toprule
\textbf{AR [\%]}  & \textbf{0} & \textbf{1} & \textbf{2} & \textbf{3} & \textbf{4} & \textbf{5}\\
\cmidrule(r){1-1}\cmidrule(r){2-2}\cmidrule(r){3-3}\cmidrule(r){4-4}\cmidrule(r){5-5}\cmidrule(r){6-6}\cmidrule(r){7-7}
\textbf{1-bit} & 5.302 / 15.228 & 74.510 / 92.048 & 75.172 / 92.500 & 75.214 / 92.482  & 75.698 / 92.656 & 75.568 / 92.662\\
\textbf{2-bit} & 65.754 / 86.718 & 75.652 / 92.658 & 75.638 / 92.702 & 75.660 / 92.512 & 75.338 / 92.660 & 75.576 / 92.615 \\
\textbf{3-bit} & 75.486 / 92.608 & 75.708 / 92.592 & {\bf 75.920 / 92.858} & 75.930 / 92.964 & 75.892 / 92.938 & 75.734 / 92.630 \\
\textbf{4-bit} & 75.700 / 92.750 & 75.784 / 92.670 & 75.880 / 92.926
 & 75.790 / 92.712 & 75.846 / 92.694 & 75.916 / 92.858\\
\midrule
\multicolumn{2}{c}{\textbf{5-bit with AR 0~\%}} & 75.600 / 92.610 &\multicolumn{2}{c}{\textbf{6-bit with AR 0~\%}} & 75.922 / 92.832 \\
\multicolumn{2}{c}{\textbf{7-bit with AR 0~\%}} &75.887 / 92.792 & \multicolumn{2}{c}{\textbf{8-bit with AR 0~\%}} & 75.670 / 92.846 \\
\bottomrule
\end{tabular}
}
\vspace{2mm}
\caption{Top-1/top-5 accuracy [\%] of ResNet-50 with various bitwidth \& AR configurations. All networks are initialized at the same condition and the network trained with full-precision activations gives the accuracy of 75.916/92.904\%.}
\label{sweep}
\end{table}

Table \ref{rvquant} shows top-1/top-5 accuracy of ResNet-50 under RV-Quant. As the table shows, RV-Quant gives similar results to V-Quant, e.g., top-1 accuracy of 3-bit 2\% RV-Quant gives an equivalent result to full precision. Compared with V-Quant, RV-Quant reduces the memory cost by 1 bit per neuron. Thus, the configuration of 3-bit 2\% RV-Quant gives 7.5X (=1/(3/32 + 0.04)) reduction in the memory cost of stored activations.
In addition, we can further reduce the memory cost of stored activations by applying traditional compression techniques to the reduced-precision activations.
In the case of 3-bit 2\% RV-Quant for ResNet-50, by applying Lempel-Ziv compression, we can further reduce the memory cost of the 3-bit data by 24.4\%, which corresponds to 9.0x reduction in the memory cost of the whole stored activations.

\begin{table}[ht]
\small
\centering
\resizebox{\columnwidth}{!}{%
\begin{tabular}{ccccccc}
\toprule
\textbf{AR [\%]}  & \textbf{0} & \textbf{1} & \textbf{2} & \textbf{3} & \textbf{4} & \textbf{5}\\
\cmidrule(r){1-1}\cmidrule(r){2-2}\cmidrule(r){3-3}\cmidrule(r){4-4}\cmidrule(r){5-5}\cmidrule(r){6-6}\cmidrule(r){7-7}
\textbf{2-bit} & 35.518 / 60.864 & 75.338 / 92.560 & 75.408 / 92.490 & 75.666 / 92.594 & 75.498 / 92.460 &  75.272 / 92.646 \\
\textbf{3-bit} & 75.156 / 92.548 & 75.876 / 92.798 & 75.932 / 92.698 & 75.658 / 92.744 & 75.906 / 92.752 & 75.488 / 92.580 \\
\bottomrule
\end{tabular}
}
\vspace{2mm}
\caption{Top-1/top-5 accuracy [\%] of ResNet-50 under RV-Quant.}
\label{rvquant}
\end{table}

Table \ref{accuracy_all} compares the accuracy of neural networks under full-precision training and two RV-Quant configurations. As the table shows, 3-bit 2\% RV-Quant gives almost the same training accuracy as full-precision training for all the networks.
% Deeper better?

\begin{table}[ht]
\small
\centering
\resizebox{\columnwidth}{!}{%
\begin{tabular}{ccccccccc}
\toprule

 & 
\textbf{AlexNet} & 
\textbf{ResNet-18} & 
\textbf{SqueezeNet-1.1} & 
\textbf{MobileNet-v2} &
\textbf{VGG-16} & 
\textbf{Inception-v3} &
\textbf{ResNet-152} & 
\textbf{DenseNet-201}\\
\cmidrule(r){1-1}\cmidrule(r){2-2}\cmidrule(r){3-3}\cmidrule(r){4-4}\cmidrule(r){5-5}\cmidrule(r){6-6}\cmidrule(r){7-7}\cmidrule(r){8-8}\cmidrule(r){9-9}
\textbf{Full} & 56.354 / 79.020 &	69.908 / 89.384 &	58.672 / 81.052 &	70.104 / 89.736 & 71.862 / 90.484 & 74.194 / 91.920 & 77.954 / 94.024 & 77.418 / 93.586 \\
\textbf{3-bit 2\%}	& 56.142 / 78.986 &	69.920 / 89.230 &	58.528 / 80.942 &	70.116 / 89.764 & 71.744 / 90.462 & 74.140 / 91.916 & 77.758 / 93.894 & 77.276 / 93.442 \\
\textbf{8-bit 0\%}&	56.238 / 78.948 &	70.010 / 89.276 &	58.750 / 81.290 &	70.294 / 89.638 & 71.774 / 90.660 & 74.224 / 92.084 & 78.354 / 93.948 & 77.320 / 93.508 \\
\bottomrule
\end{tabular}
}
\vspace{2mm}
\caption{Training results. Full means the results of conventional full-precision training, while 3-bit 2\% and 8-bit 0\% correspond to RV-Quant.}
\label{accuracy_all}
\end{table}

Table \ref{memory_cost} compares the total memory cost of activations (both stored quantized and full-precision working activations) in training with 256 mini-batch size. We compare two existing methods and three RV-Quant configurations. 'Full' represents the memory cost of conventional training with full-precision activation. As a baseline, we use the checkpointing method of Chen et al. \cite{Sublinear} since it is superior to others including \cite{Reversible}, especially for deep neural networks.
We calculate the memory cost of the checkpointing method to account for the minimum amount of intermediate activations to re-compute correct activations while having the memory cost of O($\sqrt{N}$) where $N$ is the number of layers \cite{Sublinear}.

The table shows that, compared with the checkpointing method, RV-Quant gives significant reductions in the total memory cost of activations.
For instance, in the case of ResNet-152 which is favorable to the checkpointing method due to the simple structure as well as a large number of layers, ours reduces the memory cost by 41.6\% (from 5.29GB to 3.09GB). In networks having more complex sub-networks, e.g., Inception modules, ours gives more reductions. In the case of Inception-v3, ours gives a reduction of 53.7\% (3.87GB to 1.79GB).
Note that in the case of AlexNet, the reduction is not significant. It is because the input data occupy the majority of stored activations and we store them in full precision.
However, the impact of input data storage diminishes in deep networks.

We also measured the training runtime of ResNet-50 with mini-batch of 64 on NVIDIA Tesla M40 GPU. Compared to the runtime of existing full-precision training, our method requires a small additional runtime, 8.8\% while the checkpointing method has much larger runtime overhead, 32.4\%. Note that as mentioned in Section \ref{potential}, our method has a potential of further reduction in training time on hardware platforms supporting reduced-precision computation.

\begin{table}[ht]
\small
\centering
\resizebox{\columnwidth}{!}{%
\begin{tabular}{cccccccccc}
\toprule

 & \textbf{AlexNet} & \textbf{ResNet-18} & \textbf{SqueezeNet-1.1} & 
\textbf{MobileNet-v2} & \textbf{ResNet-50} & \textbf{VGG-16} & \textbf{Inception-v3} & \textbf{ResNet-152} & \textbf{DenseNet-201} \\
\cmidrule(r){1-1}\cmidrule(r){2-2}\cmidrule(r){3-3}\cmidrule(r){4-4}\cmidrule(r){5-5}\cmidrule(r){6-6}\cmidrule(r){7-7}\cmidrule(r){8-8}\cmidrule(r){9-9}\cmidrule(r){10-10}
\thead{\textbf{Full} \\ } & 0.35 & 1.86 & 1.58 & 7.34 & 9.27 & 9.30 & 9.75 & 20.99 & 24.53 \\ 
\textbf{Chen et al. \cite{Sublinear}} & x & \thead{0.98 \\ (52.1 \%)} & \thead{1.05 \\ (66.9 \%)} & \thead{4.21 \\ (52.1 \%)}  & \thead{3.70 \\ (39.9 \%)} & x & \thead{3.87 \\ (39.8 \%)} & 
\thead{5.29 \\ (25.2 \%)} & \thead{6.62 \\ (27.0 \%)}  \\

\textbf{(2,0)} & \thead{0.23 \\ (66.4 \%)} & \thead{0.42 \\ (22.6 \%)} & \thead{0.59 \\ (37.5 \%)} & \thead{0.74 \\ (10.0 \%)} & \thead{1.22 \\ (13.2 \%)} & \thead{3.65 \\ (39.2 \%)} & \thead{1.16 \\ (11.9 \%)} & 
\thead{1.64 \\ (7.78 \%)} & \thead{2.09 \\ (8.51 \%)}  \\

\textbf{(3,0)} & \thead{0.23 \\ (67.8 \%)} & \thead{0.46 \\ (24.3 \%)} & \thead{0.61 \\ (38.8 \%)} & \thead{0.84 \\ (11.4 \%)} & \thead{1.34 \\ (14.5 \%)} & \thead{3.75 \\ (40.3 \%)} & \thead{1.43 \\ (14.8 \%)} & 
\thead{2.27 \\ (10.8 \%)} & \thead{2.85 \\ (11.6 \%)}  \\

\textbf{(3,2)} & \thead{2.40 \\ (69.5 \%)} & \thead{0.50 \\ (26.5 \%)} & \thead{0.64 \\ (40.4 \%)} & \thead{1.13 \\ (15.4 \%)} & \thead{1.52 \\ (16.4 \%)} & \thead{3.88 \\ (41.7 \%)} & \thead{1.79 \\ (18.4 \%)} & 
\thead{3.09 \\ (14.7 \%)} & \thead{3.83 \\ (15.6 \%)}  \\

\bottomrule
\end{tabular}
}
\vspace{2mm}
\caption{Comparison of memory cost (in GB).}
\label{memory_cost}
\end{table}

%Figure \ref{loss} compares the training loss of ResNet-50 between full-precision training and our proposed methods. Among our methods, the activation annealing reduces AR from 3\% to XX\% over the training period. As the figure shows, activation annealing gives almost the same loss as that of full-precision training.

Table \ref{sensitivity} shows the impact of RV-Quant configurations on training accuracy of ResNet-50.
We change the configurations when the learning rate changes (with the initial value of 0.1) at 0.01 and 0.001. For instance, (F)-(3,2)-(2,0) represents the case that, as the initial configuration, we use full-precision activation (F) during back-propagation. After 30 epochs, the configuration is changed to 3-bit 2\% RV-Quant. Then, after 60 epochs, it is changed to 2-bit 0\% RV-Quant.

\begin{table}[ht]
\small
\centering
\resizebox{\columnwidth}{!}{%
\begin{tabular}{cccccccc}
\toprule
\textbf{Configuration} & Accuracy &
\textbf{Configuration} & Accuracy  &\textbf{Configuration} & Accuracy &\textbf{Configuration} & Accuracy\\
\cmidrule(r){1-1}\cmidrule(r){2-2}\cmidrule(r){3-3}\cmidrule(r){4-4}
\cmidrule(r){5-5}\cmidrule(r){6-6}\cmidrule(r){7-7}\cmidrule(r){8-8}
\textbf{(3,2)-(2,1)-(2,0)} & 75.012 / 92.424 & \textbf{(2,0)-(2,1)-(3,2)} & 47.348 / 72.314 &

\textbf{(3,2)-(3,1)-(3,0)} & 75.720 / 92.694 & \textbf{(3,0)-(3,1)-(3,2)} & 75.604 / 92.768 \\

%\textbf{(3,4)-(3,2)-(3,0)} & 75.850 / 92.640 & \textbf{(3,0)-(3,2)-(3,4)} & 75.552 / 92.792 &

\textbf{(F)-(3,2)-(2,0)} & 75.454 / 92.628 & 
\textbf{(2,0)-(3,2)-(F)} & 50.360 / 75.024 &

\textbf{(3,2)-(3,1)-(2,0)} & 75.336 / 92.554 & 
\textbf{(2,0)-(3,1)-(3,2)} & 48.672 / 73.536  \\

\textbf{(F)-(2,1)-(2,0)} & 75.380 / 92.438 & 
\textbf{(2,0)-(2,1)-(F)} & 52.724 / 76.764 \\

\bottomrule
\end{tabular}
}
\vspace{2mm}
\caption{Sensitivity analysis of RV-Quant configurations (bitwidth and AR [\%]) across training phases}
\label{sensitivity}
\end{table}

In Table \ref{sensitivity}, the key observation is that it is important to have high precision at the beginning of training. Compared with the case that training starts with full-precision activations and ends with aggressively reduced precision, (F)-(3,2)-(2,0), the opposite case, (2,0)-(3,2)-(F) gives significantly lower accuracy, 75.454\% vs. 50.360\%. 
Another important observation is that activation annealing works. For instance, (3,2)-(3,1)-(3,0) gives almost the same result to (3,2)-(3,2)-(3,2) in Table \ref{accuracy_all} and, a more aggressive case, (3,2)-(3,1)-(2,0) gives only by 0.584\% smaller accuracy. 
Thus, as training advances, we need the smaller amount of large data, which means we can have smaller memory cost of activations. This can be exploited for memory management in servers. We expect it can also be utilized in memory-efficient server-mobile co-training in federated learning \cite{Federated} where the later stage of training requiring smaller memory cost can be performed on memory-limited mobile devices while meeting the requirements of user-specific adaptation using private data.

\begin{figure}
    \centering
    \includegraphics[width=0.6\columnwidth]{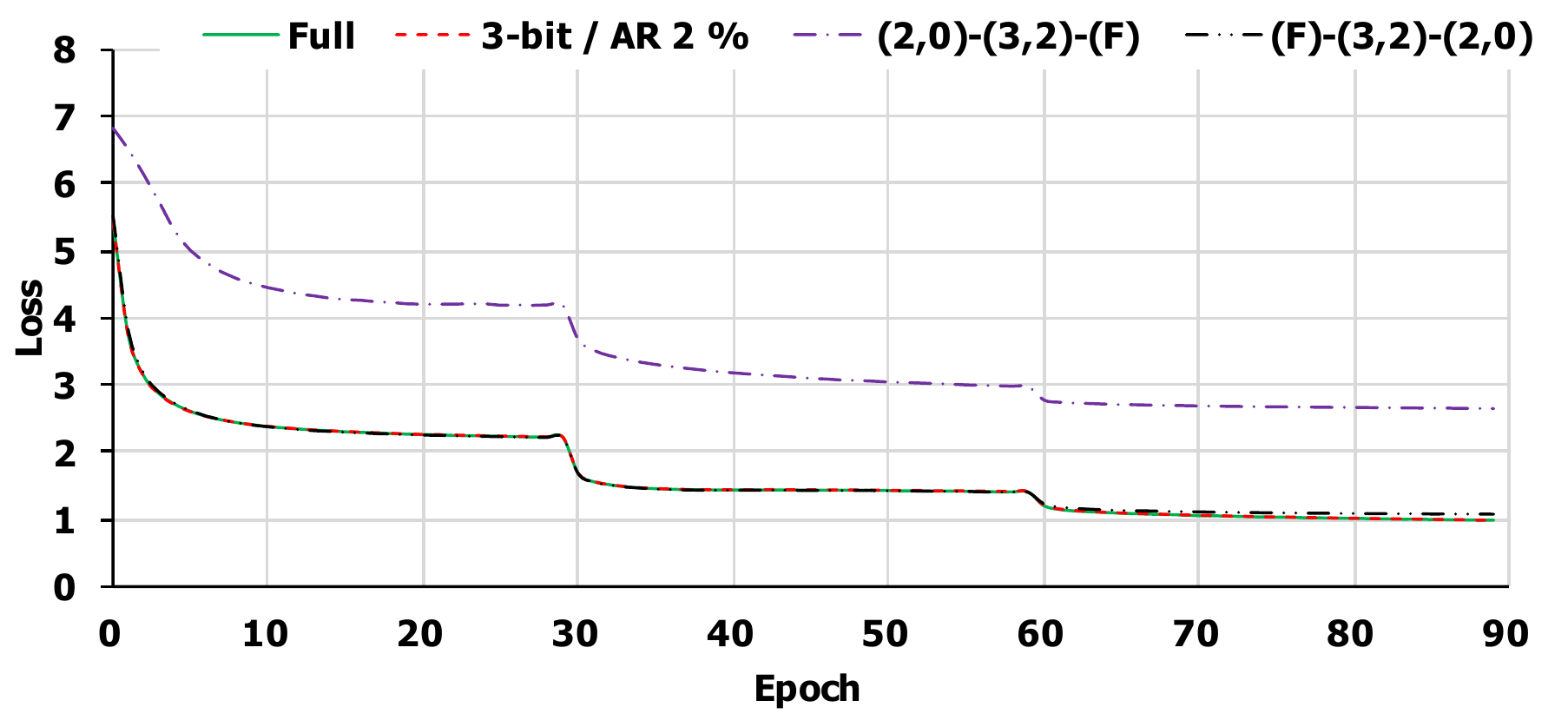}
    \caption{Training loss of ResNet-50 with various RV-Quant configurations.}
    \label{loss_curve}
\end{figure}

Figure \ref{loss_curve} shows the training loss of different RV-Quant configurations during training. First, the figure shows that too aggressive quantization in the beginning of training, i.e., (2,0)-(3,2)-(F), does not catch up with the loss of full-precision training (Full in the figure). The figure also shows that the configuration of 3-bit 2\% RV-Quant gives almost the same loss as the full-precision training.

\subsection{Inference Results}

\begin{figure}
    \centering
    \includegraphics[width=\columnwidth]{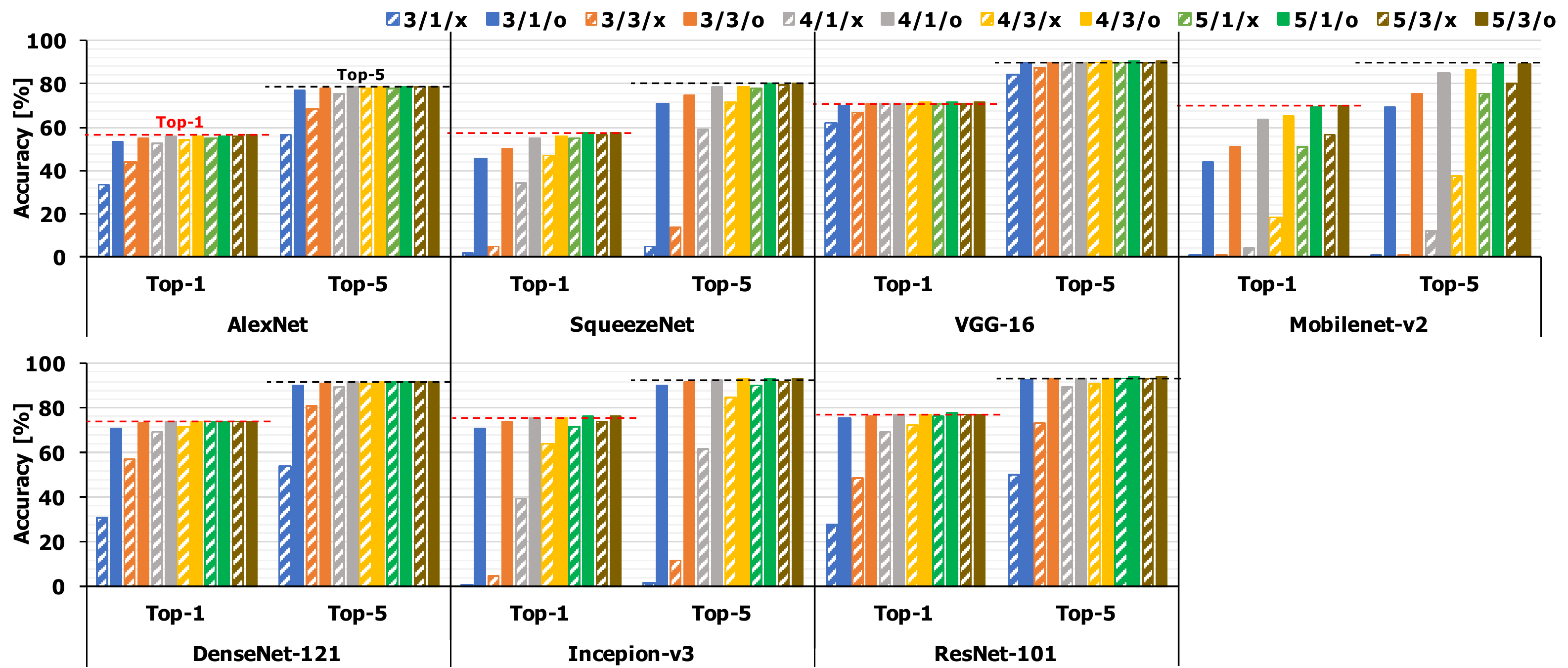}
    \caption{V-Quant results. The dashed lines represent full-precision accuracy. Legend: bitwidth/AR [\%]/fine-tuning or not.}
    \label{accuracy_inf}
\end{figure}

Figure \ref{accuracy_inf} shows the accuracy of quantized models across different configurations of bitwidth and AR. We apply the same bitwidth of low precision to both weights and activations and 16 bits to large values of weights and activations. In addition, we quantize all the layers including the first (quantized weights) and last convolutional layers. As the figure shows, V-Quant with fine-tuning, at 4 bits and an AR of 1\%, gives accuracy comparable to full precision in all the networks within 1\% of top-1 accuracy.
If V-Quant is applied without fine-tuning, the larger AR needs to be used to compensate for accuracy drop due to quantization.
However, the figure shows that fine-tuning successfully closes the accuracy gap between V-Quant and full-precision networks.

%\begin{figure}
%    \centering
%    \includegraphics[width=0.5\columnwidth]{finetuning.eps}
%    \caption{Fine-tuning vs. retraining-free results. The dashed lines represent full-precision accuracy. Legend: bitwidth/outlier ratio (\%)/fine-tuning or not.}
%    \label{fig:finetuning}
%\end{figure}

\begin{figure}
    \centering
    \includegraphics[width=0.7\columnwidth]{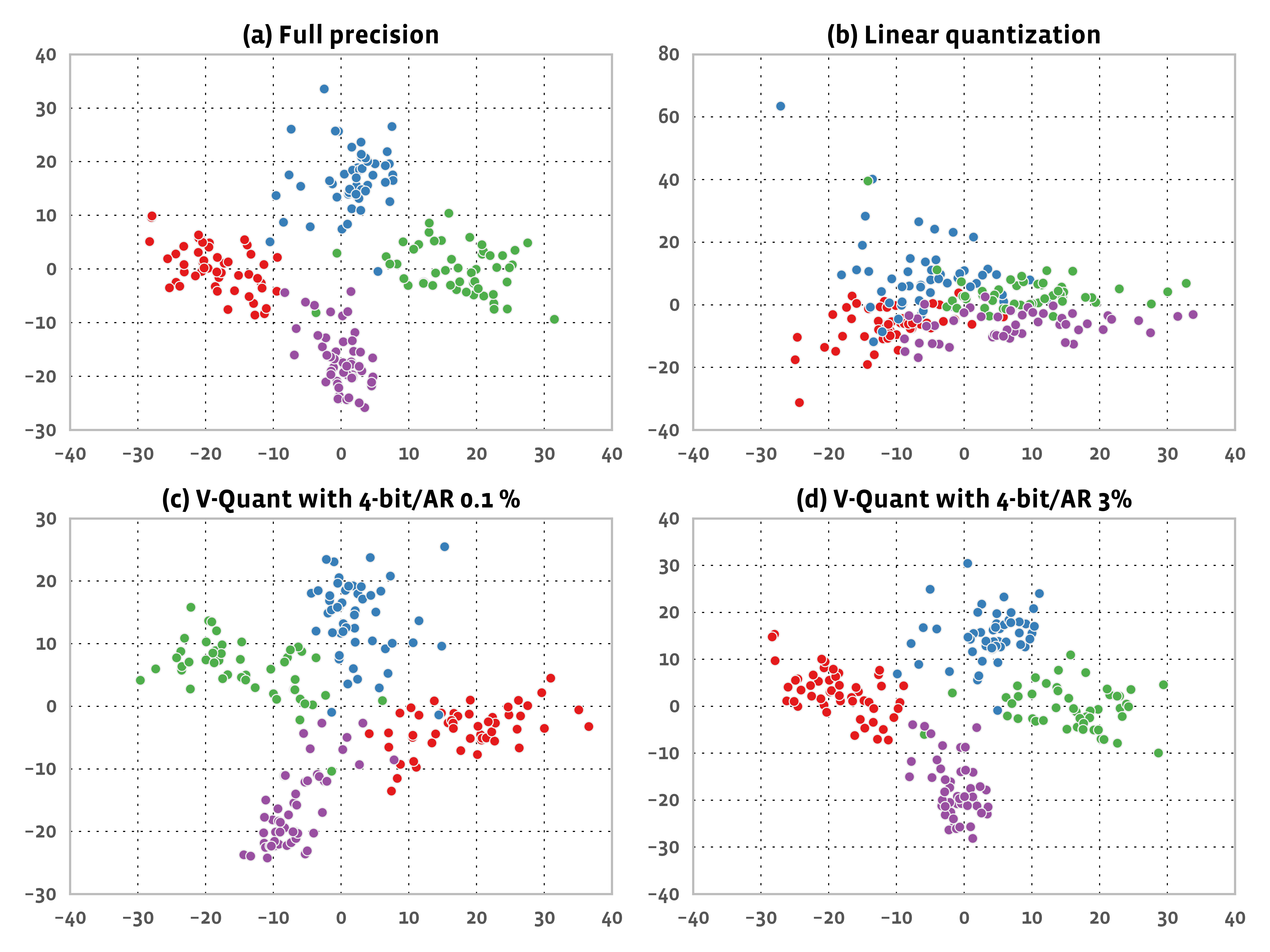}
    \caption{PCA analysis of the activations on the last convolutional layer of AlexNet.}
    \label{pca}
\end{figure}

Figure \ref{pca} illustrates the effect of large values on the classification ability. The figure shows the principal component analysis (PCA) results of the last convolutional layer of AlexNet for four classes (four colors).
Figure \ref{pca} (a) shows the PCA result of full-precision network.
As Figure \ref{pca} (b) shows, when the conventional 4-bit linear quantization, or 4-bit 0\% V-Quant is applied to weights/activations, it is difficult to successfully classify four groups of data. However, as Figure \ref{pca} (c) shows, only a very small amount (0.1\%) of large values can improve the situation. As more large values are utilized, the classification ability continues to improve (3\% in Figure \ref{pca} (d)). The figure demonstrates that our idea of reducing quantization errors for the majority of data by separately handling large data is effective in keeping good representations.

\subsection{LSTM Language Model}

We apply V-Quant to an LSTM for word-level language modeling \cite{LSTM1, LSTM2, LSTM3}. Table \ref{lstm} shows the results for the models. Each of the large and small models has two layers. The large model has 1,500 hidden units and the small one 200 units. We measure word-level perplexity on Penn Tree Bank data~\cite{ptb}. 
We apply V-Quant only to the weights of the models since clipping is applied to the activation.\footnote{The distribution of activations obtained by clipping tends to have large population near the maximum/minimum values. Considering that clipped activation functions like ReLU6 are useful, it will be interesting to further investigate clipping-aware quantization to take into account such large values.}

As Table \ref{lstm} shows, we evaluate three cases of bitwidth, 2, 3 and 4 bits and two ratios of large weights, 1\% and 3\%. 
As the table shows, for the large model, the 4-bit 1\% V-Quant preserves the accuracy of the full-precision model. However, the small model requires the larger ratio of large weights (3\%) in order to keep the accuracy.

\begin{table}[ht]
\small
\centering
\resizebox{0.5\columnwidth}{!}{%
\begin{tabular}{ccccccccc}
\toprule
               & \multicolumn{2}{c}{\textbf{Large-1\%}}  & \multicolumn{2}{c}{\textbf{Large-3\%}}  & \multicolumn{2}{c}{\textbf{Small-1\%}} & \multicolumn{2}{c}{\textbf{Small-3\%}}\\
               \cmidrule(r){2-3}\cmidrule{4-5}\cmidrule{6-7}\cmidrule(l){8-9}
               & \textbf{Valid}    & \textbf{Test}   & \textbf{Valid}    & \textbf{Test}    & \textbf{Valid}    & \textbf{Test} & \textbf{Valid}    & \textbf{Test}   \\ \midrule
\textbf{float} & 75.34             & 72.31           & 75.34             & 72.31            & 103.64            & 99.24     & 103.64     & 99.24       \\
\textbf{2-bit} & 79.92             & 77.31           & 77.87             & 74.99            & 140.70            & 135.11    & 122.25     & 117.76         \\
\textbf{3-bit} & 76.19             & 73.22           & 75.79             & 72.72            & 107.60            & 102.82    & 105.99     & 101.44         \\
\textbf{4-bit} & 75.46             & 72.48           & 75.44             & 72.44            & 104.22            & 99.83     & 103.95      & 99.57    \\
\bottomrule
\end{tabular}
}
\vspace{2mm}
\caption{Impact of quantization on word-level perplexity of an LSTM for language modeling.}
\label{lstm}

\end{table}

\section{Conclusions}
We presented a novel value-aware quantization to reduce memory cost in training and computation/memory cost in inference. In order to realize aggressively low precision, we proposed separately handling a small amount of large data and applying reduced precision to the majority of small data, which contributes to reducing total quantization errors. In order to apply our idea to training, we proposed quantized back-propagation which utilizes quantized activations only during back-propagation. For inference, we proposed applying fine-tuning to quantized networks to recover from accuracy loss due to quantization. Our experiments show that our proposed method significantly outperforms the state-of-the-art method of low-cost memory in training in deep networks, e.g., 41.6\% and 53.7\% smaller memory cost in ResNet-152 and Inception-v3, respectively. It also enables 4-bit inference (with 1\% large data) for deep networks such as ResNet-101 and DenseNet-121, and 5-bit inference for efficient networks such as SqueezeNet-1.1 and MobileNet-v2 within 1\% of additional top-1 accuracy loss.

\clearpage

\bibliographystyle{splncs03}

\end{document}